\documentclass[10pt,twocolumn,letterpaper]{article}

\usepackage{cvpr}
\usepackage{times}
\usepackage{epsfig}
\usepackage{graphicx}
\usepackage{amsmath}
\usepackage{amssymb}
\usepackage{url}

\usepackage[utf8]{inputenc}
\usepackage[T1]{fontenc}

\usepackage{booktabs,multirow}

\usepackage{enumitem}
\usepackage{placeins}
\usepackage{subfigure}

\usepackage{color}
\usepackage{authblk}

\usepackage[pagebackref=true,breaklinks=true,letterpaper=true,colorlinks,bookmarks=false]{hyperref}

\cvprfinalcopy 

\def\cvprPaperID{18} 

\ifcvprfinal\fi
\begin{document}

\def\mA{\mathcal{A}}
\def\mB{\mathcal{B}}
\def\mC{\mathcal{C}}
\def\mD{\mathcal{D}}
\def\mE{\mathcal{E}}
\def\mF{\mathcal{F}}
\def\mG{\mathcal{G}}
\def\mH{\mathcal{H}}
\def\mI{\mathcal{I}}
\def\mJ{\mathcal{J}}
\def\mK{\mathcal{K}}
\def\mL{\mathcal{L}}
\def\mM{\mathcal{M}}
\def\mN{\mathcal{N}}
\def\mO{\mathcal{O}}
\def\mP{\mathcal{P}}
\def\mQ{\mathcal{Q}}
\def\mR{\mathcal{R}}
\def\mS{\mathcal{S}}
\def\mT{\mathcal{T}}
\def\mU{\mathcal{U}}
\def\mV{\mathcal{V}}
\def\mW{\mathcal{W}}
\def\mX{\mathcal{X}}
\def\mY{\mathcal{Y}}
\def\mZ{\mathcal{Z}}

\def\1n{\mathbf{1}_n}
\def\0{\mathbf{0}}
\def\1{\mathbf{1}}

\def\A{{\bf A}}
\def\B{{\bf B}}
\def\C{{\bf C}}
\def\D{{\bf D}}
\def\E{{\bf E}}
\def\F{{\bf F}}
\def\G{{\bf G}}
\def\H{{\bf H}}
\def\I{{\bf I}}
\def\J{{\bf J}}
\def\K{{\bf K}}
\def\L{{\bf L}}
\def\M{{\bf M}}
\def\N{{\bf N}}
\def\O{{\bf O}}
\def\P{{\bf P}}
\def\Q{{\bf Q}}
\def\R{{\bf R}}
\def\S{{\bf S}}
\def\T{{\bf T}}
\def\U{{\bf U}}
\def\V{{\bf V}}
\def\W{{\bf W}}
\def\X{{\bf X}}
\def\Y{{\bf Y}}
\def\Z{{\bf Z}}

\def\a{{\bf a}}
\def\b{{\bf b}}
\def\c{{\bf c}}
\def\d{{\bf d}}
\def\e{{\bf e}}
\def\f{{\bf f}}
\def\g{{\bf g}}
\def\h{{\bf h}}
\def\i{{\bf i}}
\def\j{{\bf j}}
\def\k{{\bf k}}
\def\l{{\bf l}}
\def\m{{\bf m}}
\def\n{{\bf n}}
\def\o{{\bf o}}
\def\p{{\bf p}}
\def\q{{\bf q}}
\def\r{{\bf r}}
\def\s{{\bf s}}
\def\t{{\bf t}}
\def\u{{\bf u}}
\def\v{{\bf v}}
\def\w{{\bf w}}
\def\x{{\bf x}}
\def\y{{\bf y}}
\def\z{{\bf z}}

\def\balpha{\mbox{\boldmath{$\alpha$}}}
\def\bbeta{\mbox{\boldmath{$\beta$}}}
\def\bdelta{\mbox{\boldmath{$\delta$}}}
\def\bgamma{\mbox{\boldmath{$\gamma$}}}
\def\blambda{\mbox{\boldmath{$\lambda$}}}
\def\bsigma{\mbox{\boldmath{$\sigma$}}}
\def\btheta{\mbox{\boldmath{$\theta$}}}
\def\bomega{\mbox{\boldmath{$\omega$}}}
\def\bxi{\mbox{\boldmath{$\xi$}}}
\def\bnu{\mbox{\boldmath{$\nu$}}}                                  
\def\bphi{\mbox{\boldmath{$\phi$}}}

\def\bDelta{\mbox{\boldmath{$\Delta$}}}
\def\bOmega{\mbox{\boldmath{$\Omega$}}}
\def\bPhi{\mbox{\boldmath{$\Phi$}}}
\def\bLambda{\mbox{\boldmath{$\Lambda$}}}
\def\bSigma{\mbox{\boldmath{$\Sigma$}}}
\def\bGamma{\mbox{\boldmath{$\Gamma$}}}

\newcommand{\myminimum}[1]{\mathop{\textrm{minimum}}_{#1}}
\newcommand{\mymaximum}[1]{\mathop{\textrm{maximum}}_{#1}}    
\newcommand{\mymean}[1]{\mathop{\textrm{mean}}_{#1}}
\newcommand{\myvar}[1]{\mathop{\textrm{Variance}}_{#1}}
\newcommand{\mymin}[1]{\mathop{\textrm{minimize}}_{#1}}
\newcommand{\mymax}[1]{\mathop{\textrm{maximize}}_{#1}}
\newcommand{\mymins}[1]{\mathop{\textrm{min.}}_{#1}}
\newcommand{\mymaxs}[1]{\mathop{\textrm{max.}}_{#1}}  
\newcommand{\myargmin}[1]{\mathop{\textrm{argmin}}_{#1}} 
\newcommand{\myargmax}[1]{\mathop{\textrm{argmax}}_{#1}} 
\newcommand{\myst}{\textrm{s.t. }}

\newcommand{\denselist}{\itemsep -1pt}
\newcommand{\sparselist}{\itemsep 1pt}

\definecolor{pink}{rgb}{0.9,0.5,0.5}
\definecolor{purple}{rgb}{0.5, 0.4, 0.8}   
\definecolor{gray}{rgb}{0.3, 0.3, 0.3}
\definecolor{mygreen}{rgb}{0.2, 0.6, 0.2}

\newcommand{\cyan}[1]{\textcolor{cyan}{#1}}
\newcommand{\red}[1]{\textcolor{red}{#1}}  
\newcommand{\blue}[1]{\textcolor{blue}{#1}}
\newcommand{\magenta}[1]{\textcolor{magenta}{#1}}
\newcommand{\pink}[1]{\textcolor{pink}{#1}}
\newcommand{\green}[1]{\textcolor{green}{#1}} 
\newcommand{\gray}[1]{\textcolor{gray}{#1}}    
\newcommand{\mygreen}[1]{\textcolor{mygreen}{#1}}    
\newcommand{\purple}[1]{\textcolor{purple}{#1}}       

\definecolor{greena}{rgb}{0.4, 0.5, 0.1}
\newcommand{\greena}[1]{\textcolor{greena}{#1}}

\definecolor{bluea}{rgb}{0, 0.4, 0.6}
\newcommand{\bluea}[1]{\textcolor{bluea}{#1}}
\definecolor{reda}{rgb}{0.6, 0.2, 0.1}
\newcommand{\reda}[1]{\textcolor{reda}{#1}}

\newcommand{\mtodo}[1]{{\color{red}$\blacksquare$\textbf{[TODO: #1]}}}
\newcommand{\myheading}[1]{\vspace{1ex}\noindent \textbf{#1}}

\def\changemargin#1#2{\list{}{\rightmargin#2\leftmargin#1}\item[]}
\let\endchangemargin=\endlist
                                               
\newcommand{\cm}[1]{}

\def\xbi{\overline{\x}_i}
\def\wbi{\overline{\w}_{(i)}}
\def\wb{\overline{\w}}
\def\Ib{\overline{\I}}
\def\invC{\C^{-1}}
\def\invCi{\C_{(i)}^{-1}}
\def\ab{\overline{\balpha}}
\def\abi{\overline{\balpha}_{(i)}}
\def\Kb{\overline{\K}}
\def\Xb{\overline{\X}}
\def\kbi{\overline{\k}_{i}}
\def\Kzz{\K_{\z\z}}
\def\Kzx{\K_{\z\x}}
\def\Xsub{\X_{sub}}
\def\ssub{\s_{sub}}
\def\wbsub{\overline{\w}_{sub}}
\def\dsub{\d_{sub}}
\def\invCsub{\C^{-1}_{sub}}
\def\etal{\emph{et al}.}
\def\etals{\emph{et al}. }
\def\DS{\textcolor{red}}
\newcommand{\norm}[1]{\left\lVert#1\right\rVert}

\title{Weakly Labeling the Antarctic: The Penguin Colony Case}

\author[1,3]{Hieu Le}
\author[2,3]{Bento Gonçalves}
\author[1]{Dimitris Samaras}
\author[2,3]{Heather Lynch}
\affil[1]{Department of Computer Science, Stony Brook University}
\affil[2]{Department of Ecology and Evolution, Stony Brook University}
\affil[3]{Institute for Advanced Computational Science, Stony Brook University}

\maketitle

\begin{abstract}
Antarctic penguins are important ecological indicators -- especially in the face of climate change. In this work, we present a deep learning based model for semantic segmentation of Adélie penguin colonies in high-resolution satellite imagery. To train our segmentation models, we take advantage of the Penguin Colony Dataset: a unique dataset with 2044 georeferenced cropped images from 193 Adélie penguin colonies in Antarctica. In the face of a scarcity of pixel-level annotation masks, we propose a weakly-supervised framework to effectively learn a segmentation model from weak labels. We use a classification network to filter out data unsuitable for the segmentation network. This segmentation network is trained with a specific loss function, based on the average activation, to effectively learn from the data with the weakly-annotated labels. Our experiments show that adding weakly-annotated training examples significantly improves segmentation performance, increasing the mean Intersection-over-Union from 42.3 to 60.0\% on the Penguin Colony Dataset. 
\end{abstract}
\def\subFigSzab{\linewidth} 
\section{Introduction}

The vast and growing catalogs of high-resolution earth observation imagery present us with unprecedented opportunities for understanding ecological and geological processes, but time and domain expertise are in high demand and building a training dataset sufficient for deep learning is often not feasible. Fortunately, many earth observation applications benefit from dynamics that are slow relative to the repeat frequency of the available imagery. As a result, each image is similar to previous imagery, and prior information in the form of lower resolution or auxiliary information can be used to greatly improve classification accuracy. The use of prior knowledge naturally extends to the classification of imagery time series, which in aggregate can be used to understand the dynamics of landscape change. 
\begin{figure}[!t]
 \centering
    \includegraphics[width=0.5\textwidth]{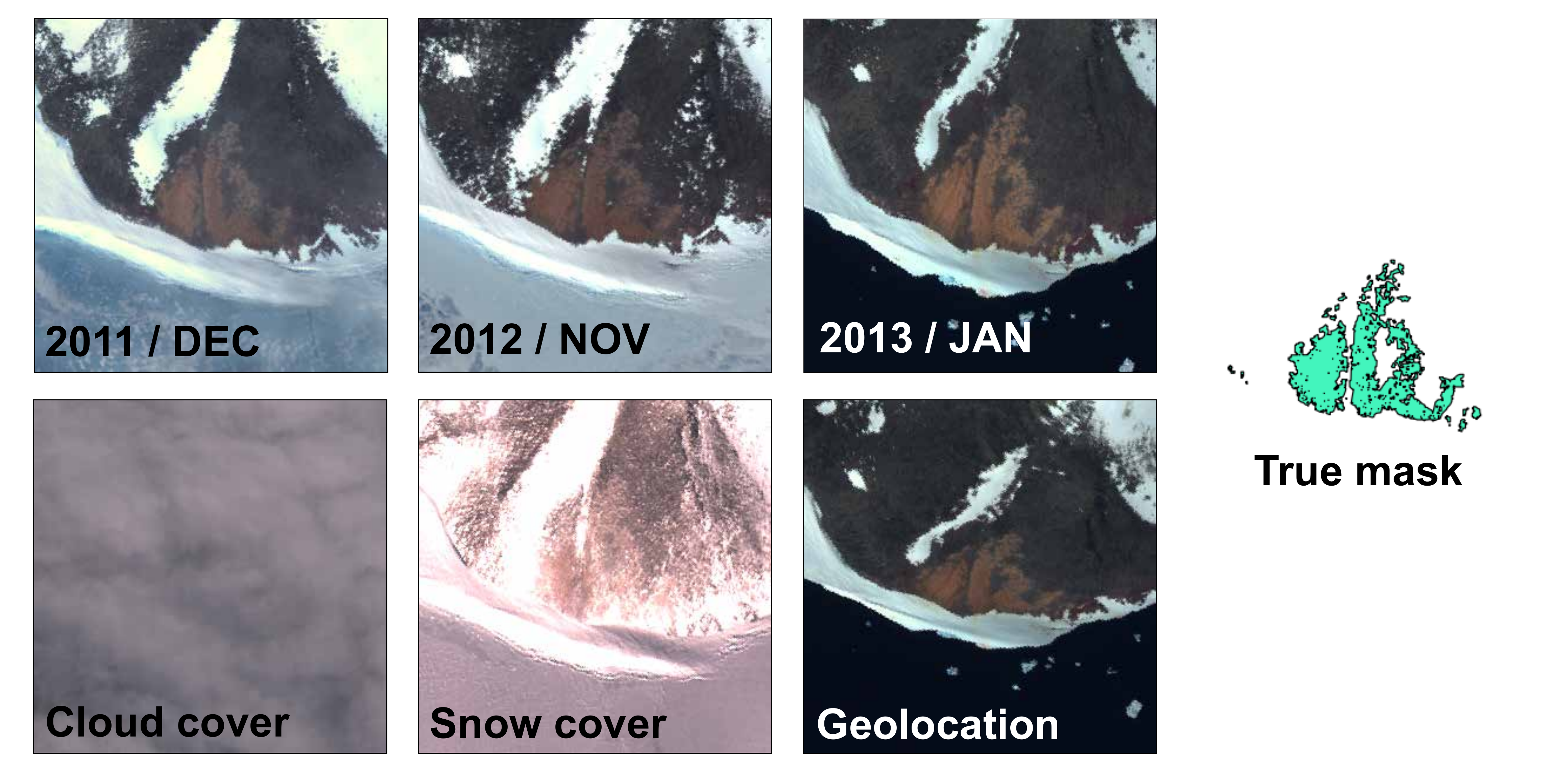}
    \vspace{0 mm}
    \caption{\textbf{Penguin colony guano mask extraction on high-resolution satellite imagery.} In the upper row, we show high-quality crops for three consecutive years of the Cape Crozier Adélie penguin colony during the breeding season. Brown to red shapes at the center of crops show the guano stains associated with the breeding colony that, when converted to area, can be used to approximate breeding population size. Boxes in the lower row show unsuitable images cropped at the same location, suffering from occlusion by clouds, heavy snow, and gross orthorectification artifacts. The goal of this work is to use weakly-annotated data, in the form of percent guano coverage, to generate better segmentation masks. Imagery copyright DigitalGlobe, Inc. 2019.
    }
    \label{fig:time-series}
    \vspace{-5mm}
\end{figure}

High-resolution satellite imagery is becoming an efficient means to survey inaccessible, or perilous, regions remotely (e.g., \cite{bjorgo2000refugee,lynch_detection_2012}). However, real-world applications for semantic segmentation often lack pixel-wise annotations because generating them is so time consuming. In this work we present a framework for weakly-supervised segmentation on georeferenced datasets. Our approach circumvents data acquisition limitations by using pixel-wise mismatched masks, unsuitable by themselves as segmentation ground-truth, to improve segmentation results. Such derived quantities are more robust to geolocation errors than pixel-wise segmentation masks and are easily acquired by a spatial query for available satellite imagery. We demonstrate our approach by segmenting penguin colonies visible in high-resolution satellite imagery, but our method is more broadly applicable to high-resolution segmentation problems common in satellite image analysis.

Antarctic penguins, being sensitive to climate-change driven shifts in the environment \cite{ainley2002adelie,forcada2009penguin} and amenable to satellite-based surveys \cite{larue2014method}, are ideal ecological indicators for the Southern Ocean ecosystem. Brush-tailed penguins, a group of three species in the genus \textit{Pygoscelis}, nest together in large colonies on snow-free rock outcrops along the Antarctic coastline. During the austral summer penguin breeding season, colonies create large-scale guano stains that are visible from space \cite{lynch_detection_2012} and with an area proportional to breeding population size \cite{larue2014method}. In addition to snapshots of population size, we can take advantage of the site-fidelity of penguins to extract time-series of population change by repeatedly surveying colonies via satellite. Time-series of guano stain shape and areal extent are invaluable to furthering our understanding of penguin population dynamics -- especially relevant in the face of climate change \cite{cimino2016projected}. 

Previous satellite-based brush-tailed penguin surveys relied on manual annotation from domain experts \cite{larue2014method,lynch_detection_2012} or, when automated, suffered from poor transferability between images \cite{witharana2016object}. Despite their limitations, satellite-based surveys have been used successfully in a variety of contexts, from finding new penguin super-colonies \cite{borowicz2018multi} to facilitating the first global population database for Adélie penguins (\textit{Pygoscelis adeliae}) \cite{lynch14first}. Manually annotating a single penguin colony, however, takes at least 30 minutes and often significantly longer. Such a laborious process, coupled with the large number of images needed to attain sufficiently temporal depth for time-series analyses at the pan-Antarctic scale, creates an urgent need for robust, automated approaches.


To amass data to train an automated guano extraction tool, we use a small number of hand-drawn geolocated guano polygons \cite{humphries_2017,lynch14first} as guides to query our high-resolution imagery for images containing penguin colonies. The two issues with this type of weakly-labeled data are: 1) This data-extraction routine is error-prone, potentially generating training images where the corresponding guano stain is not visible, and 2) Re-purposed segmentation masks are imprecise since penguin colonies keep evolving overtime and the images of the same colony are not correctly registered at the pixel level (Fig.  \ref{fig:time-series}).

In this paper, we propose a semi-supervised learning framework and a specific loss function to train a segmentation network from a small set of human-annotated images and the weakly-labeled data. We first train a classifier network, C-Net, to verify if an image contains visible guano stains. After being trained, the C-Net can be used to filter out unsuitable training images from the weakly-labeled training set, removing images that were covered in snow, shadows, clouds, or simply were not captured during the penguin breeding season. We demonstrate that, even with a small set of human-annotated images, C-Net successfully weeds out a large proportion of potentially misleading weakly-labeled training images. We employ a C-Net-filtered weakly-labeled training set, combined with our small set of human-annotated images, to train a segmentation network, S-Net, for penguin guano segmentation, using a specific loss function: For hand-labeled images, S-Net is trained to predict pixel-wise guano masks, whereas for weakly-labeled images, S-Net is trained to match percent cover. Our framework with C-Net and S-Net addresses two challenges: 1) how to determine if an image whose coordinates overlap with a penguin colony contains visible guano; and 2) how to use misaligned masks to train segmentation models.

The main contributions of this paper are three-fold: \textbf{1)} We propose a data acquisition scheme for georeferenced images, showcased with the Penguin Colony Dataset. From a few hand-drawn segmentation masks, our scheme generates thousands of weakly-labelled training images.  \textbf{2)} We propose a weakly-supervised framework using a specific loss function to learn segmentation masks from weak labels, in the form of percent cover, and a classifier network to filter out bad training examples. This approach is easily extensible to other georeferenced datasets for segmentation.\textbf{ 3)} We present test results showing that predictions from our framework are superior to pure semantic segmentation approaches and two other baselines across a range of settings. 

\section{Related Works}



The expert annotation labor required to produce segmentation masks hinders the feasibility of fully-supervised deep learning methods.
Hence, many deep learning based segmentation work focus on learning from more easily obtainable, weakly-supervised, or synthetic data \cite{Buhmann12weak,LeICCV2017,Liu2014FashionPW,Liu2013WeaklySupervisedDC,Vezhnevets2011WeaklySS,xu15weak,appleShrivastavaPTSW16,m_Le-etal-ECCV18}. A typical example of weak supervision is applying bounding boxes to learn segmentation masks \cite{dai15,papandreou15weak}.  Some methods can improve segmentation results by learning from as little as a few strokes \cite{Tang2018NormalizedCL,Vicente-etal-ECCV16} or points \cite{Maninis2018DeepEC,Russakovsky2016WhatsTP}. 
Malkin \etal \cite{malkin2018label} propose the adoption of statistical descriptors, in the form of the means and variances of low-level annotation masks, to train segmentation networks for high-resolution imagery.

\section{Penguin Colony Dataset}
\label{sec:dataset}
We present the Penguin Colony Dataset, a dataset for penguin guano segmentation on high-resolution satellite imagery. Our dataset includes a set of 31 hand-labelled guano masks from 24 Adélie penguin colonies. We also provide full metadata for images cropped from high-resolution imagery. These images include penguin colony crops from four different high-resolution optical satellites: GeoEye-1, QuickBird-2, Worldview-2 and Worldview-3. Depending on sensor, resolution for our images ranges from 2.4m/pixel (QuickBird-2) to 1.2m/pixel (Worldview-3) -- the highest available on current commercial imagery.  

\begin{figure}%
\centering
\subfigure[]{%
\label{fig:first}%
\includegraphics[width=1.4in]{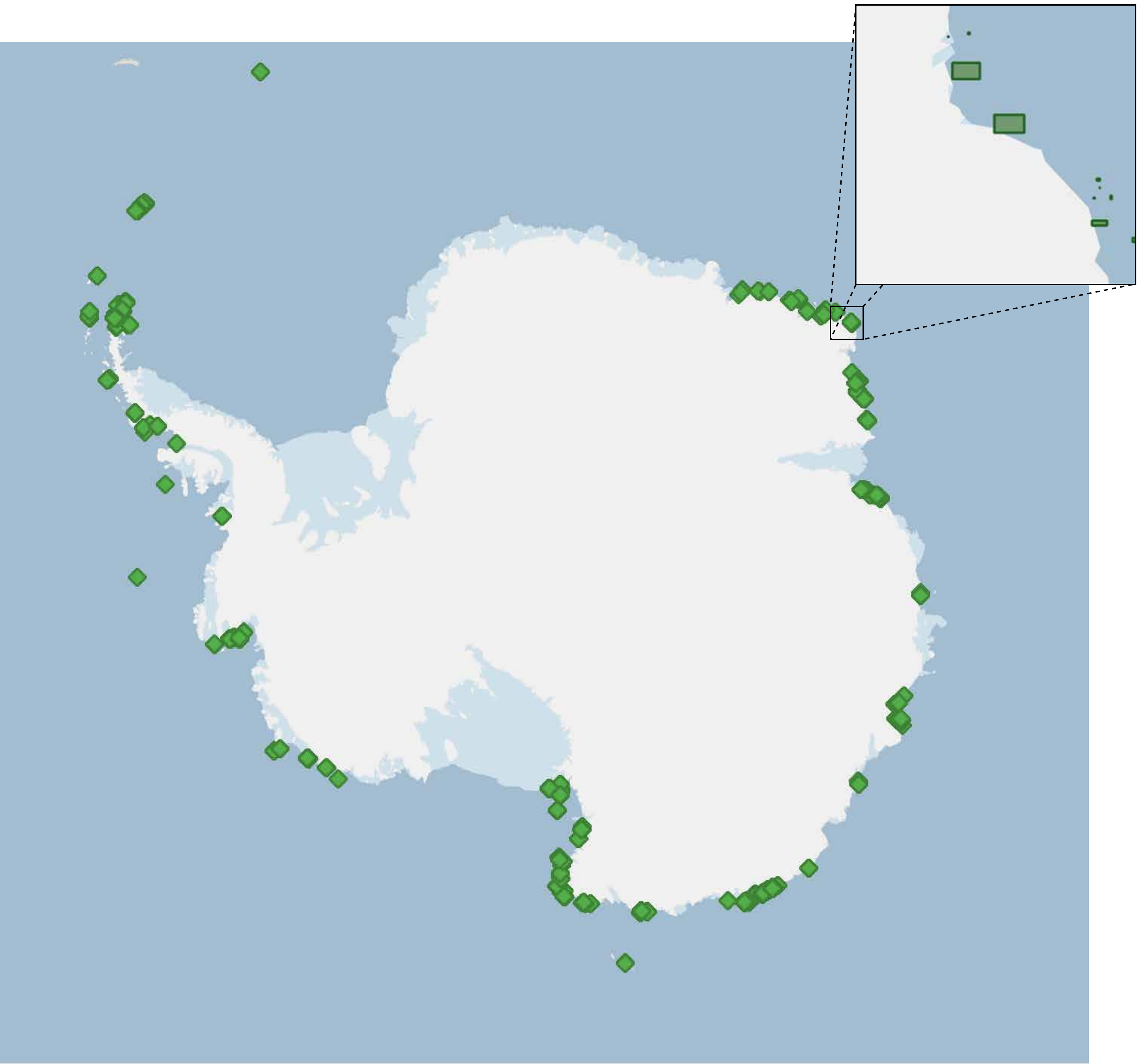}
}%
\qquad
\subfigure[]
{%
\label{fig:second}%
\includegraphics[width=1.4in]{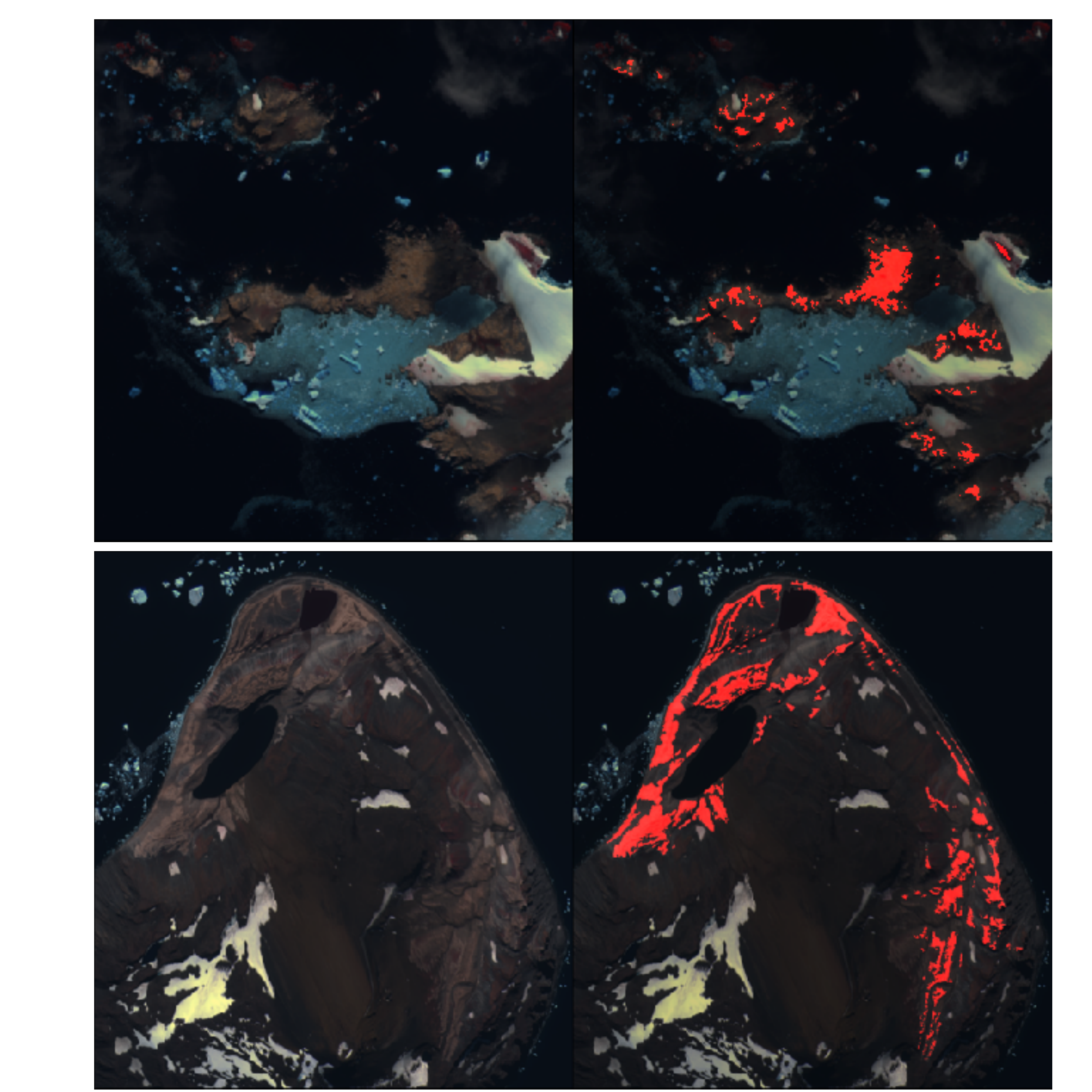}
}%
\caption{(a) Cropped image locations. Each square represents a colony bounding box (see inset) for which we found matching satellite imagery. To find matches, we query an archive of 99653 high-resolution imagery images obtained from 2002 to 2017. Our dataset harbors a total of 2044 satellite images, covering the vast majority of existing Adélie penguin colonies. (b) Two examples of hand-labelled guano mask (red overlay at right) on high-resolution imagery (left). Imagery copyright DigitalGlobe, Inc. 2019.}
    \vspace{-5mm}
\end{figure}

Adding our penguin colony polygons to medium-resolution Landsat-based masks from \cite{lynch14first}, we store the locations of 193 Adélie penguin colonies. With colony polygons in hand, we query an archive of 99653 high-resolution satellite images from the Antarctic coastline for images that encircle penguin colony shapes. We then crop each image to the smallest bounding box for each penguin colony, adding 100 pixels of padding on each side. For each cropped image generated this way, we calculate a Shannon entropy index, discarding crops that score 5 or lower. Following this automated data acquisition routine, we cropped 2044 images at locations shown in Fig. \ref{fig:first}, heretofore referenced as the "weakly-annotated dataset". These 2044 images can be grouped into a video segmentation dataset ~\cite{Khoreva15cvpr,Perazzi2017,le_accv2016_vs,yu2015} consisting of image time series for each of the 193 penguin colonies in our dataset. 

We then split the images from our 31 high-resolution masks into 18 training and 13 testing images. Similar to the weakly-annotated dataset, cropped images vary in size depending on the extent of the colony and the sensor resolution. Crops from high-resolution images for which we created segmentation masks are heretofore referenced as the "manually-labeled dataset".

In summary, we provide a dataset containing shapefiles for guano polygon masks and colony bounding boxes, and cropped images for manually-labelled and weakly-annotated penguin colonies. The weakly-annotated component of our dataset is easily expanded as our imagery archive grows. Though aircraft-based aerial imagery for related problems do exist (e.g. \cite{NOAAFish14:online}), to the best of our knowledge, this is the first public dataset involving animal population estimation from high-resolution satellite imagery. More details can be found at: \href{https://github.com/lynch-lab/CVPR19-CV4GC-WeaklyLabeling}{github.com/lynch-lab/CVPR19-CV4GC-WeaklyLabeling}

\section{Weakly-Supervised Learning for Penguin Colony Segmentation}

\begin{figure*}[!ht]
 \centering
    \includegraphics[width=0.78\textwidth]{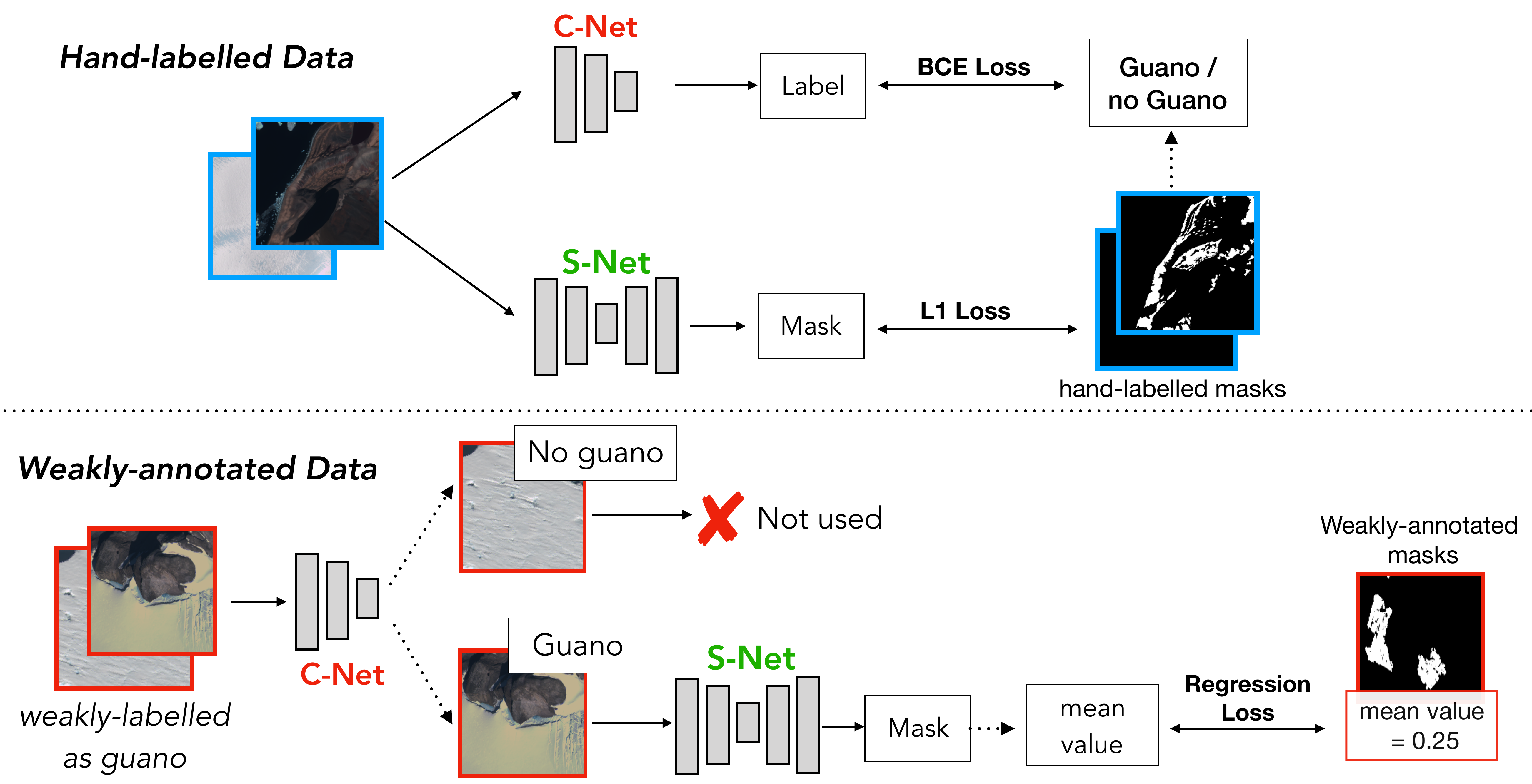}
    \vspace{1 mm}
    \caption{\textbf{Weakly-Supervised Learning Framework For Penguin Colony Segmentation.} Data with hand-labelled annotations are used to train both the C-Net and S-Net: The C-Net learns to predict the image label (if there is any penguin guano in the image) while the S-Net learns to segment the penguin guano areas. Once the C-Net is trained, it filters out images weakly-marked as containing guano but without visible guano due to snow, shadows, cloud, or poor timing relative to the breeding season. The S-Net learns from the weakly-annotated images to output the segmentation masks such that the mean activation of pixels in the predicted masks approximate the weakly-annotated masks. Imagery copyright DigitalGlobe, Inc. 2019. 
    }
    \label{fig:framework}
    \vspace{-5mm}
\end{figure*}

As discussed in Section \ref{sec:dataset}, only 0.8\% (18 images) of our training data is hand-labelled, and there are 2044 penguin colony images with misaligned segmentation masks. There are two main challenges in this scenario: The first is that the image-level labels are unavoidably noisy. The images, although captured at the locations of known penguin colonies, might not contain visible penguin guano. The images could be covered in snow, shadows, or clouds, or were not captured in the breeding season when the penguin guano is visible. The second issue is that the pixel-level annotations are misaligned with the actual image contents due to georegistration errors or orthorectification artifacts.

We propose a method to learn from those weakly-labeled data for image segmentation. We use two networks, C-Net and S-Net, to maximize learning from the hand-labelled data. C-Net is a classification network that learns to predict the image labels, e.g., whether an image contains \textbf{\textit{any}} penguin guano and S-Net is a segmentation network that learns to segment the penguin guano areas. The main purpose of the C-Net is to filter out bad training examples from the weakly-annotated training set to better train the S-Net. Our framework is summarized in Fig. \ref{fig:framework}.

The C-Net first learns to classify images from the hand-annotated data. The training label for each image is binary: \textbf{0} implies the image is without any guano and \textbf{1} otherwise. 
Once the C-Net has been trained, we use it to assist the training of the S-Net. We train the S-Net on both the hand-labelled and the weakly-annotated data. For the hand-labelled data, the S-Net is trained to predict the segmentation mask from the input image. For each weakly-annotated image, we want to mitigate the risk of using bad training examples. Hence, we use the C-Net to classify all images that are weakly-labelled as containing guano and then remove all images that are classified as "no guano" from the training pool for the S-Net.
 
As minor georegistration errors or orthorectification artifacts create mismatches between annotation masks and input images, we do not use generated crops as groundtruth segmentation masks for S-Net. Instead, we train S-Net to recover the mean pixel values on weakly-annotated masks. Such a metric works as a proxy for fractional guano coverage in the images, which is more robust to imperfect georegistration. We essentially enforce an image-level statistic matching between predicted masks and weakly-annotated masks instead of minimizing pixel-wise differences. 

 Let $I$ denote an input image, and $M(I)$ be the guano mask of $I$. Let $S(I)$ denote the output of the S-Net for the input image $I$. Ideally, the output should be~\textbf{1} for guano pixels and \textbf{0} otherwise. The objective of S-Net's training is to minimize a weighted combination of two losses:
     \vspace{-5mm}

\begin{align*}
\mathcal{L}_S(I) =
  &\lambda_{seg}\mathcal{H}(I)\norm{ S(I) - M(I) }_2 +\\ &\lambda_{reg}(1-\mathcal{H}(I))\norm{\textrm{mean}(S(I))-\textrm{mean}(M(I))}_1 
\end{align*}
where the value of $\mathcal{H}(I)$ is 1 if $I$ is a hand-labelled image and 0 if $I$ is a weakly annotated image. $\lambda_{seg}$ and $\lambda_{reg}$ control how much the S-Net should learn from the two losses respectively. We empirically set $(\lambda_{seg},\lambda_{reg})$ to $(1,5)$.
    \vspace{-5mm}


\section{Experiments}
We evaluate the performance of the C-Net and S-Net on the testing set of the Penguin Colony Dataset. The testing set contains 13 images of various sizes. To evaluate the performance of the C-Net, we crop the testing images into patches of size $256\times256$ with a step size of 64. The label for each image patch is obtained from the corresponding cropped mask.
To train the S-Net, we crop each training image to small patches of size $384\times384$ with a step size of $192$ to reduce I/O bottleneck issues arising due to the large sizes of images. From the original training set, we obtain 6055 hand-labelled and 100584 weakly-annotated training patches.
To evaluate the performance of the S-Net, we compare the output of the S-Net to the hand-annotated guano masks of the Penguin Colony testing set.

\begin{table}[]
\caption{\textbf{C-Net classification results on the Penguin Colony Dataset.} Confusion matrices summarizing the results of C-Net on the image patches cropped from the Penguin Colony Dataset testing set (a) and on image patches cropped from the weakly-annotated set (b; W.A. Set). ``Guano'' patches contain penguin guano areas and ``No Guano'' patches contain only background. ``\textit{True Label}'' means the patches are manually annotated. ``\textit{Weak Label}'' means the labels are obtained via the possibly mismatched masks. An image patch is classified as positive if C-Net outputs a positive score and negative otherwise.}
\label{tab:C-Net1}
\setlength{\tabcolsep}{12pt}
\vspace{3pt}
\begin{tabular}{llll}
\multicolumn{2}{c}{\textbf{(a) Testing Set}}   & \multicolumn{2}{c}{\textit{True Label}}                              \\ \cline{3-4} 
                                                 & \multicolumn{1}{l|}{}      & \multicolumn{1}{l|}{Guano} & \multicolumn{1}{l|}{No Guano} \\ \cline{2-4} 
\multicolumn{1}{l|}{\multirow{2}{*}{\textit{Pred.}}} & \multicolumn{1}{l|}{Guano}  & \multicolumn{1}{l|}{858}  & \multicolumn{1}{l|}{469}   \\ \cline{2-4} 
\multicolumn{1}{l|}{}                            & \multicolumn{1}{l|}{No Guano} & \multicolumn{1}{l|}{265}   & \multicolumn{1}{l|}{13968}  \\ \cline{2-4} 
\end{tabular}\\
\vspace{1pt}

\begin{tabular}{llll}
\multicolumn{2}{c}{\textbf{(b) W.A. Set}}  & \multicolumn{2}{c}{\textit{Weak Label}}                              \\ \cline{3-4} 
                                                 & \multicolumn{1}{l|}{}      & \multicolumn{1}{l|}{Guano} & \multicolumn{1}{l|}{No Guano} \\ \cline{2-4} 
\multicolumn{1}{l|}{\multirow{2}{*}{\textit{Pred.}}} & \multicolumn{1}{l|}{Guano}  & \multicolumn{1}{l|}{9597}  & \multicolumn{1}{l|}{2070}   \\ \cline{2-4} 
\multicolumn{1}{l|}{}                            & \multicolumn{1}{l|}{No Guano} & \multicolumn{1}{l|}{19446}   & \multicolumn{1}{l|}{69471}  \\ \cline{2-4} 
\end{tabular}
\end{table}

We design the C-Net based on Resnet-18 \cite{Xie_2017_CVPR} and the S-Net based on U-Net \cite{unet15a}. We use stochastic gradient descent with the Adam solver~\cite{Adam} to train our models. 

\subsection{Penguin Guano Classification.}

We first analyze the classification performance of the C-Net. 
We evaluate the C-Net on the testing set consisting of 15560 patches of sizes $256\times256$, which are cropped from 13 testing images with a step size of 64. Table \ref{tab:C-Net1}(a) reports the confusion matrix summarizing the result of the C-Net on the testing image patches. The C-Net achieves a 0.65 precision and 0.76 recall on this set. For the weakly-annotated image patches shown in Table \ref{tab:C-Net1}(b), C-Net classifies 19446 patches weakly-labelled as ``Guano'' to be non-guano. These patches then are not used for training the S-Net. 
We show the performances of S-Net trained with and without these removed training patches in Section \ref{sec:S-Neteval}.

\def\subboxsize{0.12\textwidth}
\def\subfig{0.4\textwidth}
\begin{figure}[t]
 \centering
       \makebox[\subboxsize]{Input}
    \makebox[\subboxsize]{S-Net}
    \makebox[\subboxsize]{C+S-Net}\\
    \includegraphics[width=\subfig,height=1in]{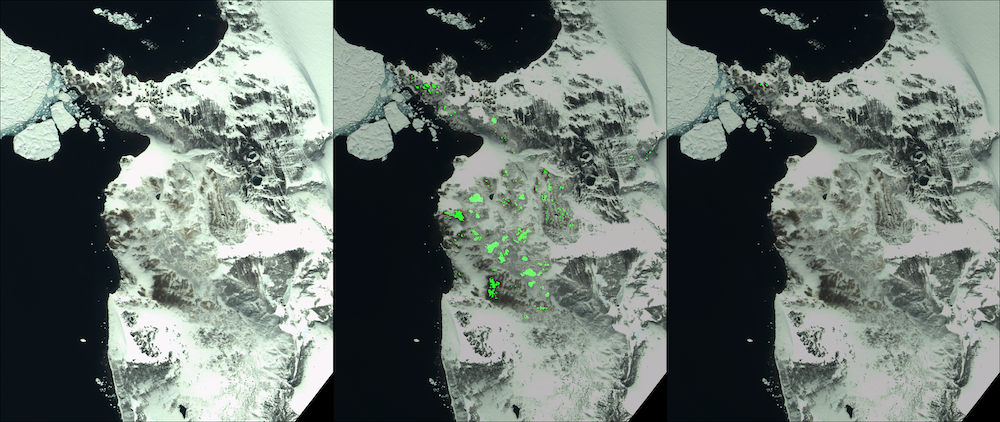}
     \includegraphics[width=\subfig]{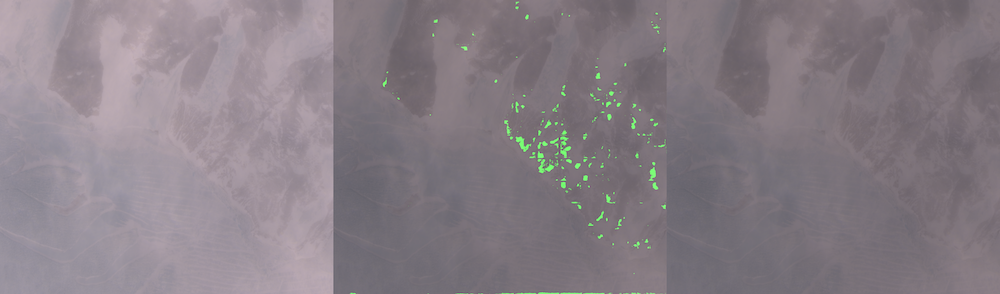}
     \caption{\textbf{The Effect of C-Net: Segmentation results of the S-Net trained on the whole dataset and on the dataset filtered by C-Net.} Input image (left column) covered by snow (top row) and fog (second row). Because the C-Net filters out noisy training examples for the S-Net, S-Net does not predict any guano pixels in either the snow or fog corrupted images. Imagery copyright DigitalGlobe, Inc. 2019.
    }
    \label{fig:C-Net}
\end{figure}

\subsection{Penguin Guano Segmentation.}
\label{sec:S-Neteval}
\setlength{\tabcolsep}{4pt}
\begin{table}[!t]
\centering
\caption{{\bf Segmentation results of our models on the Penguin Colony Dataset.} We train our S-Net on different sets of training data and loss functions. ``H'' denotes hand-labelled data and ``W'' denotes weakly-annotated data. For ``Seg. (H)'', we compute the segmentation loss on the hand-labelled data to train the network. For ``Reg. (W)'', we compute the regression loss on the weakly-annotated data. ``S.'' only uses the segmentation loss while ``SR.'' uses the weighted combination loss of the segmentation and regression losses. ``+'' uses weakly-annotated data, and ``C+S-Net'' is our S-Net trained on the C-Net filtered data.
}
\begin{tabular}{lccc}
\toprule
Method   &Data& Loss Function  & mIoU (\%)    \\
\midrule
S-Net S.  & H & Seg.(H) & 42.3\\
S-Net S. + & H+W &Seg.(H) + Seg.(W)&37.7 \\ 
S-Net SR. +& H+W & Seg.(H) + Reg.(W) \  & 55.0   \\
C+S-Net SR. + & H+W & Seg.(H) + Reg.(W) \  &  \textbf{60.0}  \\
\bottomrule
\end{tabular}
\label{tab:final}
    \vspace{-1mm}
\end{table}

We evaluate our penguin colony segmentation network on the Penguin Colony Dataset. To obtain the output segmentation mask for an image, we first crop the image into patches of size $256\times256$ with a step size of 128. 
Each patch is input into the network to obtain a patch prediction mask. We obtain the final prediction mask for the input image by averaging all overlapped patch predictions at each pixel. We use mean Intersection-Over-Union (mIoU) to evaluate the segmentation masks.

\def\subboxsize{0.16\subFigSzab}
\def\subfig{0.8\textwidth}
\begin{figure*}[t]
 \centering
    \makebox[\subboxsize]{Input}
    \makebox[\subboxsize]{Overlaid GT}
    \makebox[\subboxsize]{S-Net}
    \makebox[\subboxsize]{S-Net}
    \makebox[\subboxsize]{C+S-Net}\\
    \makebox[\subboxsize]{}
    \makebox[\subboxsize]{}
    \makebox[\subboxsize]{Seg.(H)}
    \makebox[\subboxsize]{Seg.(H) + Reg.(W)}
    \makebox[\subboxsize]{Seg.(H) + Reg.(W)}
     \includegraphics[width=\subfig]{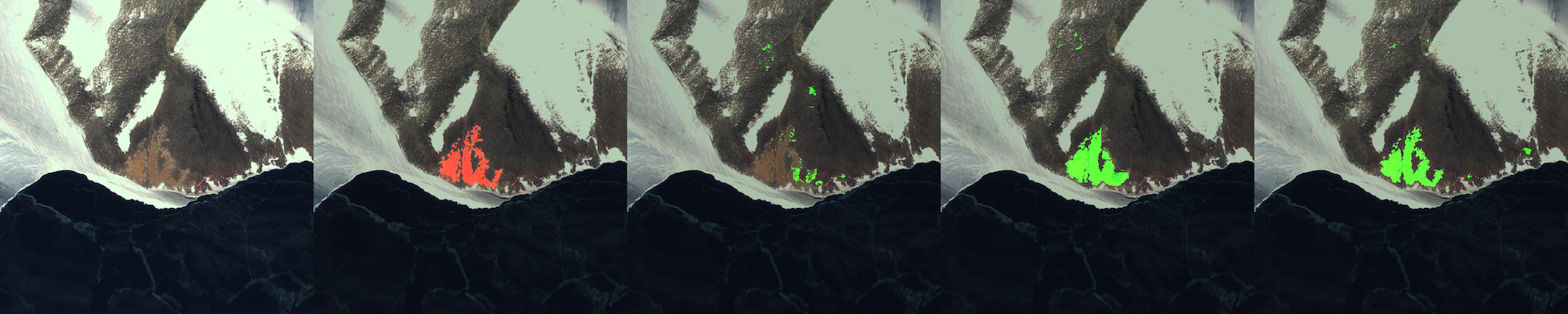}
    \includegraphics[width=\subfig]{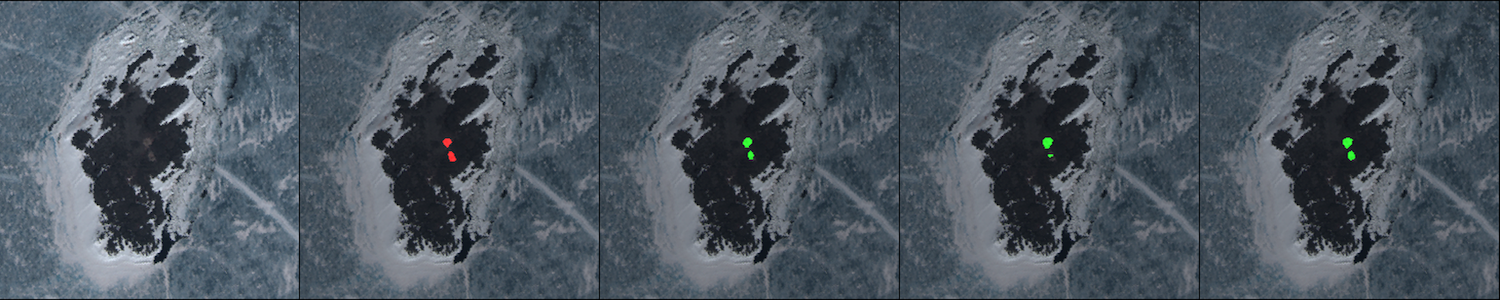}
    \includegraphics[width=\subfig]{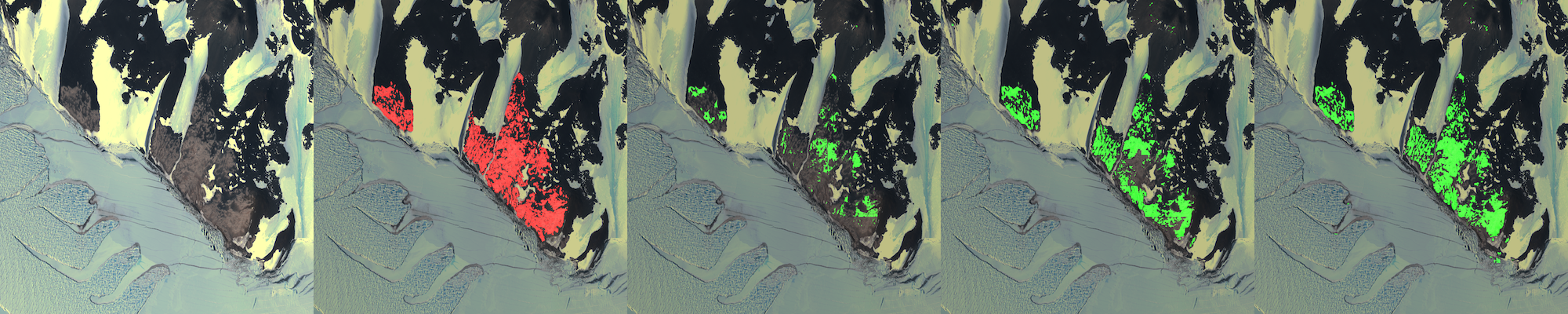}
     
     \caption{\textbf{Qualitative comparison between our method and other baseline methods on the Penguin Colony Dataset.}  All methods use S-Net as the backbone segmentation network. From left to right: the input image, the input image overlaid by the ground truth penguin guano polygon, the results of S-Net trained only with the correctly-annotated set, the results of S-Net trained on the correctly-annotated set with the segmentation loss and on the weakly-annotated set with the regression loss. The last column is our proposed method using the C-Net to filter out bad training examples for training the S-Net on both correctly-annotated and weakly-annotated set. Imagery copyright DigitalGlobe, Inc. 2019.
    }
    \label{fig:qual}
        \vspace{-3mm}
\end{figure*}

Table \ref{tab:final} summarizes the results of our model. We compare our method with three baselines. All methods use as the backbone segmentation network the network with the same architecture as the S-Net. 
The first row shows the results of the S-Net trained on only the hand-labelled data, using the segmentation loss, denoted as ``S-Net S.''.
This model achieves 42.3 mIoU on the testing set of the Penguin Colony Dataset. A straightforward use of the weakly-annotated data is to train a segmentation network to output the segmentation masks, regardless of the mis-georegistration. This network is trained using the same segmentation loss function, but with more data, denoted as ``S-Net S. +'' in the second row.
This model does not take into account the mis-georegistration issue of the guano polygons. Unsurprisingly, segmentation performance decreases from 42.3 to 37.7 mIoU score since the model is guided to output the guano pixels at the mismatched locations. 

The third row shows the effect of our mean activation regression loss to learn from the misaligned guano masks. We train a S-Net on both the hand-annotated data and weakly-annotated data. For the weakly-annotated data, this S-Net is only constrained to output the segmentation masks that have the same average pixel values as the weakly-annotated guano masks. This model is denoted as ``S-Net SR. +'' in the third row of Table \ref{tab:final}.  This simple modification improves the mIoU by 30\%, from 42.3 to 55 mIoU, compared to the model trained only on the hand-labelled training set. 

We then evaluate the effect of the C-Net.
We use C-Net to classify all 29043 training patches that are marked as containing guano, according to the weakly-annotated guano masks. We then remove image patches classified as ``No Guano'' from the training pool.
As can be seen from Table \ref{tab:C-Net1}, C-Net removes 19446 image patches, which are 67\% of the whole set of positive training patches that are weakly-annotated. With less noisy training images, S-Net achieves better segmentation performance on the Penguin Colony Dataset where the mIoU improves from 55\% to 60\% mIoU score. Fig. \ref{fig:C-Net} illustrates the effect of the C-Net. The figure shows that S-Net trained on the complete data is forced to predict the guano even on the snow covered areas while the model that trained on the filtered data, ``C+S-Net'', does not predict any guano pixels in these cases.

Fig. \ref{fig:qual} shows a qualitative comparison between our proposed method and the other three baselines discussed above. The three penguin colonies shown in the figure are (from top to bottom) Arthurson Ridge, Balaena Islands, and Cape Crozier. As can be seen, the model trained on only the hand-labelled data does not segment Arthurson Ridge and Cape Crozier correctly. The model trained without the C-Net picks up some non-guano pixels at Arthurson Ridge while missing some guano areas at Balaena Islands and Cape Crozier. The model trained on the data filtered by C-Net outputs cleaner and more complete segmentation masks.

\begin{figure}[!ht]
 \centering
    \includegraphics[width=0.43\textwidth]{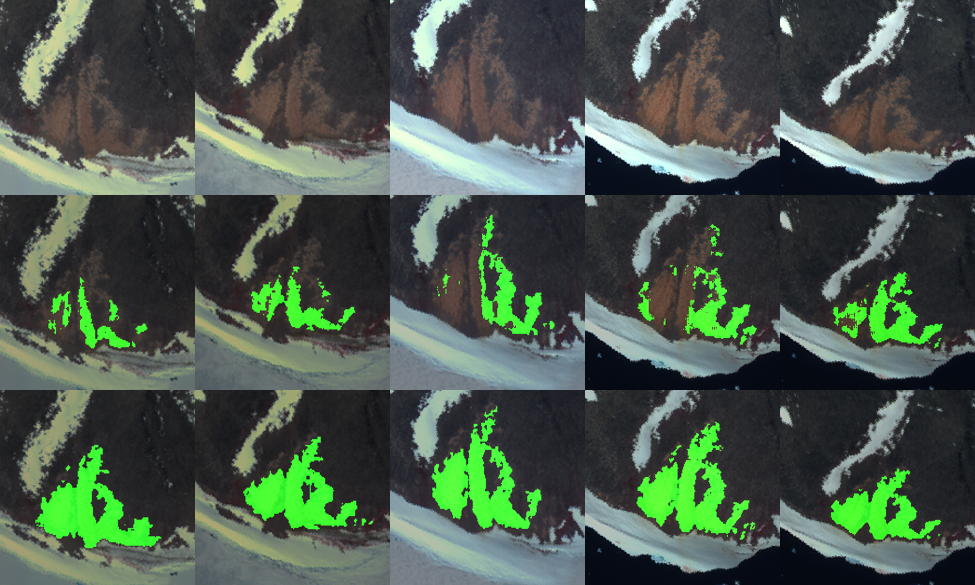}
    \vspace{1 mm}
    \caption{\textbf{Qualitative comparison of S-Net trained with and without weakly-labeled data.} 
    The top row shows the input images, the middle row visualizes the results of S-Net trained only on the hand-labelled data, and the bottom row visualizes the results of S-Net trained on the hand-labelled data and weakly-annotated data. We used our trained C-Net to filter the training data before training this S-Net. Imagery copyright DigitalGlobe, Inc. 2019.
    }
    \label{fig:Arthuson}
    \vspace{-5mm}
\end{figure}

The effect of using weakly-labelled data is shown best in Fig. \ref{fig:Arthuson}. We compare the results of our model trained with and without the weakly-annotated data for images captured at the penguin site named Arthurson Ridge throughout multiple years. The weakly-annotated data improves significantly the generalizability of the network.

\section{Conclusion}
In this work, we present an approach to easily acquire weakly-annotated data on semantic segmentation datasets with georeferenced imagery. Our data acquisition routine circumvents limitations caused by unfavorable weather conditions and bad georegistration. We successfully weed out unsuitable training data using a CNN classifier, obtaining improved segmentation performance when using a subset of 9597 positive crops selected by our classifier, instead of training with all 29043 positive crops. Employing weakly-labeled data -- curated by our CNN classifier -- dramatically improves the accuracy of our segmentation output by 42\%, increasing the mIoU from 42.3 to 60.0\%. Apart from segmenting penguin guano, our framework is easily extensible to other georeferenced segmentation applications where we wish to track changes in features with predictable location. 

\myheading{Acknowledgements.} We gratefully acknowledge assistance from the US National Science Foundation (Award No. NSF/ OPP-1255058, NSF/ ICER-1740595, NFS/ CNS-1718014), the National Geographic/Microsoft AI for Earth program, assistance from the Polar Geospatial Center, a gift from Adobe, the Partner University Fund, and the SUNY2020 Infrastructure Transportation Security Center. We would also like to acknowledge Stony Brook's Institute of Advanced Computational Science, the SeaWulf computing cluster and support staff, and Marcel Simon and Siyu Yang for helpful suggestions.

\FloatBarrier

{\small
\bibliographystyle{ieee}
\bibliography{longstrings,egbib}

\begin{thebibliography}{10}\itemsep=-1pt

\bibitem{NOAAFish14:online}
Noaa fisheries steller sea lion population count | kaggle.
\newblock www.kaggle.com/c/noaa-fisheries-steller-
  sea-lion-population-count/data.

\bibitem{ainley2002adelie}
D.~Ainley.
\newblock {\em The Ad{\'e}lie penguin: bellwether of climate change}.
\newblock Columbia University Press, 2002.

\bibitem{bjorgo2000refugee}
E.~Bjorgo.
\newblock Refugee camp mapping using very high spatial resolution satellite
  sensor images.
\newblock {\em Geocarto International}, 15(2):79--88, 2000.

\bibitem{borowicz2018multi}
A.~Borowicz, P.~McDowall, C.~Youngflesh, T.~Sayre-McCord, G.~Clucas, R.~Herman,
  S.~Forrest, M.~Rider, M.~Schwaller, T.~Hart, et~al.
\newblock Multi-modal survey of ad{\'e}lie penguin mega-colonies reveals the
  danger islands as a seabird hotspot.
\newblock {\em Scientific reports}, 8(1):3926, 2018.

\bibitem{Buhmann12weak}
J.~M. Buhmann.
\newblock Weakly supervised structured output learning for semantic
  segmentation.
\newblock In {\em Proceedings of the IEEE Conference on Computer Vision and
  Pattern Recognition}, CVPR '12, 2012.

\bibitem{cimino2016projected}
M.~A. Cimino, H.~J. Lynch, V.~S. Saba, and M.~J. Oliver.
\newblock Projected asymmetric response of ad{\'e}lie penguins to antarctic
  climate change.
\newblock {\em Scientific reports}, 6:28785, 2016.

\bibitem{dai15}
J.~Dai, K.~He, and J.~Sun.
\newblock Boxsup: Exploiting bounding boxes to supervise convolutional networks
  for semantic segmentation.
\newblock abs/1503.01640, 2015.

\bibitem{forcada2009penguin}
J.~Forcada and P.~N. Trathan.
\newblock Penguin responses to climate change in the southern ocean.
\newblock {\em Global Change Biology}, 15(7):1618--1630, 2009.

\bibitem{humphries_2017}
G.~Humphries, R.~Naveen, M.~Schwaller, C.~Che-Castaldo, P.~McDowall,
  M.~Schrimpf, and H.~Lynch.
\newblock Mapping application for penguin populations and projected dynamics
  (mapppd): data and tools for dynamic management and decision support.
\newblock {\em Polar Record}, 53(2):160–166, 2017.

\bibitem{Khoreva15cvpr}
A.~Khoreva, F.~Galasso, M.~Hein, and B.~Schiele.
\newblock Classifier based graph construction for video segmentation.
\newblock In {\em IEEE Conference on Computer Vision and Pattern Recognition
  (CVPR 2015)}, pages 951--960, Boston, MA USA, 2015. IEEE Computer Society.

\bibitem{Adam}
D.~P. Kingma and J.~Ba.
\newblock Adam: {A} method for stochastic optimization.
\newblock In {\em Proceedings of the International Conference on Learning
  Representations}, 2015.

\bibitem{larue2014method}
M.~A. LaRue, H.~Lynch, P.~Lyver, K.~Barton, D.~Ainley, A.~Pollard, W.~Fraser,
  and G.~Ballard.
\newblock A method for estimating colony sizes of ad{\'e}lie penguins using
  remote sensing imagery.
\newblock {\em Polar Biology}, 37(4):507--517, 2014.

\bibitem{le_accv2016_vs}
H.~Le, V.~Nguyen, C.-P. Yu, and D.~Samaras.
\newblock Geodesic distance histogram feature for video segmentation.
\newblock In {\em Proceedings of the Asian Conference on Computer Vision}, nov
  2016.

\bibitem{m_Le-etal-ECCV18}
H.~Le, T.~F.~Y. Vicente, V.~Nguyen, M.~H. Nguyen, and D.~Samaras.
\newblock {A+D Net}: Training a shadow detector with adversarial shadow
  attenuation.
\newblock In {\em Proceedings of the European Conference on Computer Vision},
  2018.

\bibitem{LeICCV2017}
H.~Le, C.-P. Yu, G.~Zelinsky, and D.~Samaras.
\newblock Co-localization with category-consistent features and geodesic
  distance propagation.
\newblock In {\em ICCV 2017 Workshop on CEFRL: Compact and Efficient Feature
  Representation and Learning in Computer Vision}, 2017.

\bibitem{Liu2014FashionPW}
S.~Liu, J.~Feng, C.~Domokos, H.~Xu, J.~Huang, Z.~Hu, and S.~Yan.
\newblock Fashion parsing with weak color-category labels.
\newblock {\em IEEE Transactions on Multimedia}, 16:253--265, 2014.

\bibitem{Liu2013WeaklySupervisedDC}
Y.~Liu, Z.~Li, J.~Tang, and H.~Lu.
\newblock Weakly-supervised dual clustering for image semantic segmentation.
\newblock {\em 2013 IEEE Conference on Computer Vision and Pattern
  Recognition}, pages 2075--2082, 2013.

\bibitem{lynch14first}
H.~Lynch and M.~LaRue.
\newblock First global census of the adélie penguin.
\newblock {\em The Auk}, 131(4):457--466, 2014.

\bibitem{lynch_detection_2012}
H.~J. Lynch, R.~White, A.~D. Black, and R.~Naveen.
\newblock Detection, differentiation, and abundance estimation of penguin
  species by high-resolution satellite imagery.
\newblock {\em Polar Biology}, 35(6):963--968, February 2012.

\bibitem{malkin2018label}
K.~Malkin, C.~Robinson, L.~Hou, and N.~Jojic.
\newblock Label super-resolution networks.
\newblock In {\em International Conference on Learning Representations}, 2019.

\bibitem{Maninis2018DeepEC}
K.-K. Maninis, S.~Caelles, J.~Pont-Tuset, and L.~V. Gool.
\newblock Deep extreme cut: From extreme points to object segmentation.
\newblock {\em 2018 IEEE/CVF Conference on Computer Vision and Pattern
  Recognition}, pages 616--625, 2018.

\bibitem{papandreou15weak}
G.~Papandreou, L.-C. Chen, K.~Murphy, and A.~L. Yuille.
\newblock Weakly- and semi-supervised learning of a dcnn for semantic image
  segmentation.
\newblock {\em arxiv}, 2015.

\bibitem{Perazzi2017}
F.~Perazzi, A.~Khoreva, R.~Benenson, B.~Schiele, and A.Sorkine-Hornung.
\newblock Learning video object segmentation from static images.
\newblock In {\em Computer Vision and Pattern Recognition}, 2017.

\bibitem{unet15a}
O.~Ronneberger, P.Fischer, and T.~Brox.
\newblock U-net: Convolutional networks for biomedical image segmentation.
\newblock In {\em Proceedings of the International Conference on Medical Image
  Computing and Computer Assisted Intervention}, volume 9351 of {\em LNCS},
  pages 234--241, 2015.

\bibitem{Russakovsky2016WhatsTP}
O.~Russakovsky, A.~L. Bearman, V.~Ferrari, and L.~Fei-Fei.
\newblock What's the point: Semantic segmentation with point supervision.
\newblock In {\em ECCV}, 2016.

\bibitem{appleShrivastavaPTSW16}
A.~Shrivastava, T.~Pfister, O.~Tuzel, J.~Susskind, W.~Wang, and R.~Webb.
\newblock Learning from simulated and unsupervised images through adversarial
  training.
\newblock In {\em Proceedings of the IEEE Conference on Computer Vision and
  Pattern Recognition}, 2016.

\bibitem{Tang2018NormalizedCL}
M.~Tang, A.~Djelouah, F.~Perazzi, Y.~Boykov, and C.~Schroers.
\newblock Normalized cut loss for weakly-supervised cnn segmentation.
\newblock {\em 2018 IEEE/CVF Conference on Computer Vision and Pattern
  Recognition}, pages 1818--1827, 2018.

\bibitem{Vezhnevets2011WeaklySS}
A.~Vezhnevets, V.~Ferrari, and J.~M. Buhmann.
\newblock Weakly supervised semantic segmentation with a multi-image model.
\newblock {\em 2011 International Conference on Computer Vision}, pages
  643--650, 2011.

\bibitem{Vicente-etal-ECCV16}
T.~F.~Y. Vicente, L.~Hou, C.-P. Yu, M.~Hoai, and D.~Samaras.
\newblock Large-scale training of shadow detectors with noisily-annotated
  shadow examples.
\newblock In {\em Proceedings of the European Conference on Computer Vision},
  2016.

\bibitem{witharana2016object}
C.~Witharana and H.~Lynch.
\newblock An object-based image analysis approach for detecting penguin guano
  in very high spatial resolution satellite images.
\newblock {\em Remote Sensing}, 8(5):375, 2016.

\bibitem{Xie_2017_CVPR}
S.~Xie, R.~Girshick, P.~Dollar, Z.~Tu, and K.~He.
\newblock Aggregated residual transformations for deep neural networks.
\newblock In {\em The IEEE Conference on Computer Vision and Pattern
  Recognition (CVPR)}, July 2017.

\bibitem{xu15weak}
J.~{Xu}, A.~G. {Schwing}, and R.~{Urtasun}.
\newblock Learning to segment under various forms of weak supervision.
\newblock In {\em Proceedings of the IEEE Conference on Computer Vision and
  Pattern Recognition}, June 2015.

\bibitem{yu2015}
C.-P. Yu, H.~Le, G.~Zelinsky, and D.~Samaras.
\newblock Efficient video segmentation using parametric graph partitioning.
\newblock In {\em ICCV}, 2015.

\end{thebibliography}
}

\end{document}